\documentclass[lettersize,journal]{IEEEtran}
\usepackage{amsmath,amsfonts}
\usepackage{algorithmic}
\usepackage{algorithm}
\usepackage{array}
\usepackage[caption=false,font=normalsize,labelfont=sf,textfont=sf]{subfig}
\usepackage{textcomp}
\usepackage{stfloats}
\usepackage{url}
\usepackage{verbatim}
\usepackage{graphicx}
\usepackage{cite}
\usepackage{booktabs}
\usepackage{array}
\usepackage{bm}
\usepackage{makecell}
\usepackage{booktabs}
\usepackage{multirow}
\usepackage{tikz}
\usepackage[table]{xcolor}
\usepackage{eso-pic}
\usepackage{xcolor}

\usetikzlibrary{shadows}
\usepackage[most]{tcolorbox}
\usepackage{stfloats}
\usepackage[hidelinks]{hyperref}

\begin{document}

\title{A Large-Language-Model Supported Personalized Driving Framework for Lane Change in Highway Scenarios

\author{
Dong Bi,
Yongqi Zhao,
Paul Kovacevic,
Tomislav Mihalj,
Ji Zhou,
Jiayuan Gong,
and Arno Eichberger,~\IEEEmembership{Member,~IEEE}
}

        % <-this % stops a space
\thanks{This work has been submitted to the IEEE for possible publication.
Copyright may be transferred without notice, after which this version may
no longer be accessible.~(\textit{Corresponding author: Yongqi Zhao)}}% <-this % stops a space

\thanks{Dong Bi is with the School of Intelligent Connected Vehicle, Hubei University of Automotive Technology, 442002 Shiyan and Institute of Automotive Engineering, Graz University of Technology, Graz 8010, Austria (e-mail: dong.bi@tugraz.at)}
\thanks{Yongqi Zhao, Paul Kovacevic, Tomislav Mihalj, Ji Zhou, and Arno Eichberger are with the Institute of Automotive Engineering, Graz University of Technology, Graz 8010, Austria (e-mail: yongqi.zhao@tugraz.at; paul.kovacevic@tugraz.at; tomislav.mihalj@tugraz.at; ji.zhou@student.tugraz.at; arno.eichberger@tugraz.at)}
\thanks{Jiayuan Gong is with the School of Intelligent Connected Vehicle, Hubei University of Automotive Technology, 442002 Shiyan (e-mail: jygong@huat.edu.cn)}
}

% The paper headers
\markboth{Journal of \LaTeX\ Class Files,~Vol.~14, No.~8, August~2021}%
{Shell \MakeLowercase{\textit{et al.}}: A Sample Article Using IEEEtran.cls for IEEE Journals}

%\IEEEpubid{0000--0000/00\$00.00~\copyright~2021 IEEE}
% Remember, if you use this you must call \IEEEpubidadjcol in the second
% column for its text to clear the IEEEpubid mark.

\maketitle

\begin{abstract}
Personalized driving can improve the user acceptance of automated driving systems. However, existing methods still provide limited support for translating natural-language driving preferences, especially when such preferences are expressed implicitly, into executable and distinguishable driving behaviors. This paper proposes a large language model (LLM)-supported personalized driving framework for highway lane-change scenarios. The framework maps natural-language driving commands to executable planning parameters in the open-source Apollo automated driving stack according to three driving styles: aggressive, normal, and conservative. To establish this mapping, candidate planning parameters are evaluated based on the resulting lane-change behaviors, and style-specific parameter sets are constructed through clustering and style-intensity ranking. For command interpretation, a retrieval dataset is constructed to support retrieval-augmented generation (RAG), enabling LLM-based interpretation of implicit user commands. Experimental results show that the derived parameter sets generate distinguishable personalized lane-change behaviors, while RAG consistently improves preference interpretation, particularly for implicit commands. These results indicate the potential of integrating LLM-based natural-language interaction with Apollo to support personalized lane-change behavior generation. The source code and the relevant datasets are available at: \url{https://github.com/ftgTUGraz/LLM-Personalized-Driving}.
\end{abstract}

\begin{IEEEkeywords}
Automated driving, personalized lane change, LLMs, retrieval-augmented generation, natural-language interaction, virtual testing.
\end{IEEEkeywords}

\section{Introduction}
\IEEEPARstart{F}{uture} automated vehicles are expected to move beyond safety-oriented automation toward more flexible, intelligent, and interactive mobility solutions~\cite{10529622}. Personalized driving technologies have attracted increasing attention for their potential to improve user acceptance, comfort, and driving experience in automated driving systems (ADSs)~\cite{8918081,yi2019implicit,liao2024review}. In personalized automated driving, a key problem is how to translate user preferences into vehicle behaviors. This is particularly challenging for lane changes~\cite{bevly2016lane}, where vehicle behavior depends on coupled longitudinal and lateral control and interactions with surrounding traffic participants. Therefore, personalized lane-change adaptation through natural language requires reliable interpretation of user preferences and generation of executable, distinguishable lane-change behaviors for different driving styles.

Previous works have explored user-preference modeling based on historical driving data~\cite{schrum2024maveric}, predefined driving-style labels~\cite{butakov2014personalized,yang2021personalized}, and manually designed criteria~\cite{huang2021personalized}. These studies demonstrate the feasibility of adapting vehicle behaviors to individual preferences, but they provide limited support for direct human-vehicle interaction through natural language. Recent large language model (LLM)-based automated driving studies have introduced language reasoning into high-level task planning~\cite{ma2024learning}, system reconfiguration~\cite{song2025autoware}, human-like interaction~\cite{cui2024drive}, decision support~\cite{wen2024dilu}, and style-aware trajectory generation~\cite{gao2026stylevla,kou2025padriver}. However, the use of natural-language interaction to generate distinguishable and adjustable personalized lane-change behaviors within an automated driving stack remains insufficiently explored.

Another unresolved issue is the interpretation of implicit driving preferences, where users do not directly specify a driving style but express the desired lane-change behavior indirectly. Such preferences may be conveyed through urgency, comfort needs, safety concerns, or traffic situations, rather than explicit labels such as aggressive, normal, or conservative, and are common in human–vehicle interaction~\cite{hasenjager2019survey}. However, existing LLM-based driving studies offer limited support for reliably mapping implicit preferences to executable, style-specific driving behaviors. 

In summary, prior studies still present three limitations. First, limited support is available for intuitive and flexible driving-style adaptation through natural-language interaction, especially when users express explicit preferences for aggressive, normal, or conservative driving styles. Second, the generation of distinguishable and adjustable personalized lane-change behaviors within an open-source automated driving stack remains insufficiently explored. Third, existing methods provide limited support for mapping implicit driving preferences to executable, style-specific lane-change behaviors.

To address these limitations, this paper proposes an LLM-based personalized driving framework for lane-change adaptation in highway scenarios. The framework is implemented in the open-source Apollo automated driving stack~\cite{apolloauto_apollo}. The main contributions are summarized as follows:
\begin{enumerate}
    \item A personalized lane-change framework is proposed to translate natural-language driving commands into executable lane-change behaviors through driving-style interpretation and planning-parameter mapping.
    \item Lane-change parameter sets are constructed for different driving styles to generate distinguishable behaviors and support multiple intensity levels for aggressive, normal, and conservative driving.
    \item A natural-language preference dataset is constructed for retrieval-augmented generation (RAG), supporting more accurate interpretation of explicit and implicit commands.  
\end{enumerate}

\section{RELATED WORK}
\subsection{Personalized Driving and Lane-Change Adaptation in ADS}
Existing surveys indicate that adapting ADS to driver preferences can improve user experience, whereas generic ADS may neglect individual characteristics. Personalization has been explored in representative ADS functions, including forward collision warning~\cite{xie2024personalized}, adaptive cruise control~\cite{wang2021personalized,koglbauer2017drivers}, automated driving speed adaptation~\cite{delmas2024personalizing}, data-driven driving style learning~\cite{schrum2024maveric}, and user-driven adaptation to dynamic preferences~\cite{zhang2024user}. The personalization mechanisms of these studies mainly rely on historical behavior data, predefined preference representations, or explicit feedback, rather than direct natural-language preference expression.

As a representative and challenging ADS maneuver, lane change provides an important scenario for studying personalized driving behavior~\cite{bevly2016lane}. Existing studies have modeled personalized lane-change responses~\cite{butakov2014personalized}, selected trajectories according to user preferences on safety, comfort, and stability~\cite{huang2021personalized}, and generated human-like lane-change trajectories considering driver characteristics and traffic-environmental factors~\cite{yang2021personalized}. Hu et al.~\cite{hu2025accelerating} further used user takeover interventions to update perceived safe driving zones and personalize lane-change trajectory planning online. However, these methods do not focus on natural-language-based preference expression, nor do they explicitly address how to generate distinguishable and adjustable lane-change behaviors for different driving style within a mature ADS stack.

\begin{table*}[!b]
\caption{Comparison of Related Studies on Personalized Driving and LLM-Based Language Interaction}
\label{tab:related_work_summary}
\centering
\footnotesize
\setlength{\heavyrulewidth}{1.4pt}
\setlength{\lightrulewidth}{0.5pt}
\renewcommand{\arraystretch}{1.12}
\setlength{\tabcolsep}{4pt}

\begin{tabular*}{0.98\textwidth}{
@{\extracolsep{\fill}}
>{\centering\arraybackslash}p{1.4cm}
>{\centering\arraybackslash}p{2.5cm}
>{\centering\arraybackslash}p{2.6cm}
>{\centering\arraybackslash}p{2.3cm}
>{\centering\arraybackslash}p{2.9cm}
>{\centering\arraybackslash}p{3.0cm}
}
\toprule
Work &
Personalized Behavior &
Personalized Driving Style &
Lane-Change Focus &
LLM-Based Language Interaction &
Implicit Preference Dataset \\
\midrule
{[6]} & \checkmark & \checkmark & -- & -- & -- \\
{[8]} & \checkmark & Partial & \checkmark & -- & -- \\
{[9]} & \checkmark & \checkmark & \checkmark & -- & -- \\
{[10]} & \checkmark  & -- & -- & \checkmark & -- \\
{[11]} & \checkmark & -- & -- & \checkmark & -- \\
{[12], [26]} & \checkmark & -- & -- & \checkmark & -- \\
{[21]} & \checkmark & -- & -- & -- & -- \\
{[22]} & \checkmark & -- & \checkmark & -- & -- \\
{[28]} & \checkmark & -- & -- & \checkmark & -- \\
{[36]} & \checkmark & -- & -- & \checkmark & -- \\
\textbf{Ours} &
\textbf{\checkmark} &
\textbf{\checkmark} &
\textbf{\checkmark} &
\textbf{\checkmark} &
\textbf{\checkmark} \\
\bottomrule
\end{tabular*}
\vspace{1mm}
\footnotesize
\end{table*}

\subsection{LLM-Based Language Interaction for ADS}
According to~\cite{cui2026llm4ad,yang2023llm4drive,zhou2024vision,zhao2026survey}, recent advances in LLMs have created new opportunities for human-centered automated driving by enabling natural-language understanding, intent interpretation, and high-level decision support. Cui et al.~\cite{cui2024drive,cui2024personalized,song2025autoware} developed LLM-based interaction frameworks to translate verbal commands into executable control programs. Xu et al.~\cite{xu2025personalizing} used LLMs to personalize warning messages and human-machine interaction, while Ma et al.~\cite{ma2024learning} proposed an LLM-based programming planner that converts natural-language instructions and driving contexts into executable policies in CARLA~\cite{dosovitskiy2017carla}. These studies mainly focus on command execution, communication, or general policy generation, rather than mapping driver preference expressions to executable and distinguishable driving-style behaviors.

Recent multimodal large language
models and vision-language-action models have further used to explore personalized or style-conditioned driving behavior generation. StyleVLA~\cite{gao2026stylevla} fine-tunes a vision-language-action model to generate physically plausible trajectories under predefined driving-style instructions, such as comfort, sporty, and safety. PADriver~\cite{kou2025padriver} enables switching among slow, normal, and fast driving modes through predefined personalized prompts and selects discrete driving actions according to the traffic environment. In addition, Ge et al.~\cite{ge2024llm} proposed an LLM-based operating-system architecture for task understanding, module coordination, and system management. These studies mainly focus on predefined style instructions, instruction-driven ADS adaptation, or system-level coordination, while the interpretation of open-ended, implicit user preferences into personalized driving styles remains insufficiently explored.

\subsection{Natural-Language Understanding for Driving Preference Interpretation}
To bridge the gap between natural human preference expression and machine-interpretable driving systems, recent studies have explored natural-language understanding in ADS. Yang et al.~\cite{yang2024human} used LLMs to infer structured system requirements from in-cabin verbal commands, while Liao et al.~\cite{liao2024gpt} improved command grounding by combining textual, emotional, visual, and contextual cues. Dataset-oriented works such as doScenes~\cite{roy2025doscenes}, Talk2Car~\cite{deruyttere2019talk2car}, and NuPrompt~\cite{wu2025language} connect natural-language instructions or prompts with real driving scenes, objects, and trajectories. LMDrive~\cite{shao2024lmdrive} further incorporates natural-language instructions into closed-loop end-to-end driving in CARLA. Although these works demonstrate the potential of natural language for driving-scene understanding and behavior generation, limited attention has been paid to interpreting implicit preferences across different personalized driving styles.

In summary, existing studies have explored personalized driving, LLM-based driving interaction, and natural-language scene understanding from different perspectives. However, as summarized in~\autoref{tab:related_work_summary}, translating natural-language preferences into executable personalized driving styles within a ADS framework remains insufficiently studied. In particular, implicit expressions are challenging to interpret.

\section{METHODOLOGY}
\begin{figure*}[!t]
\centering
\includegraphics[width=7.0in]{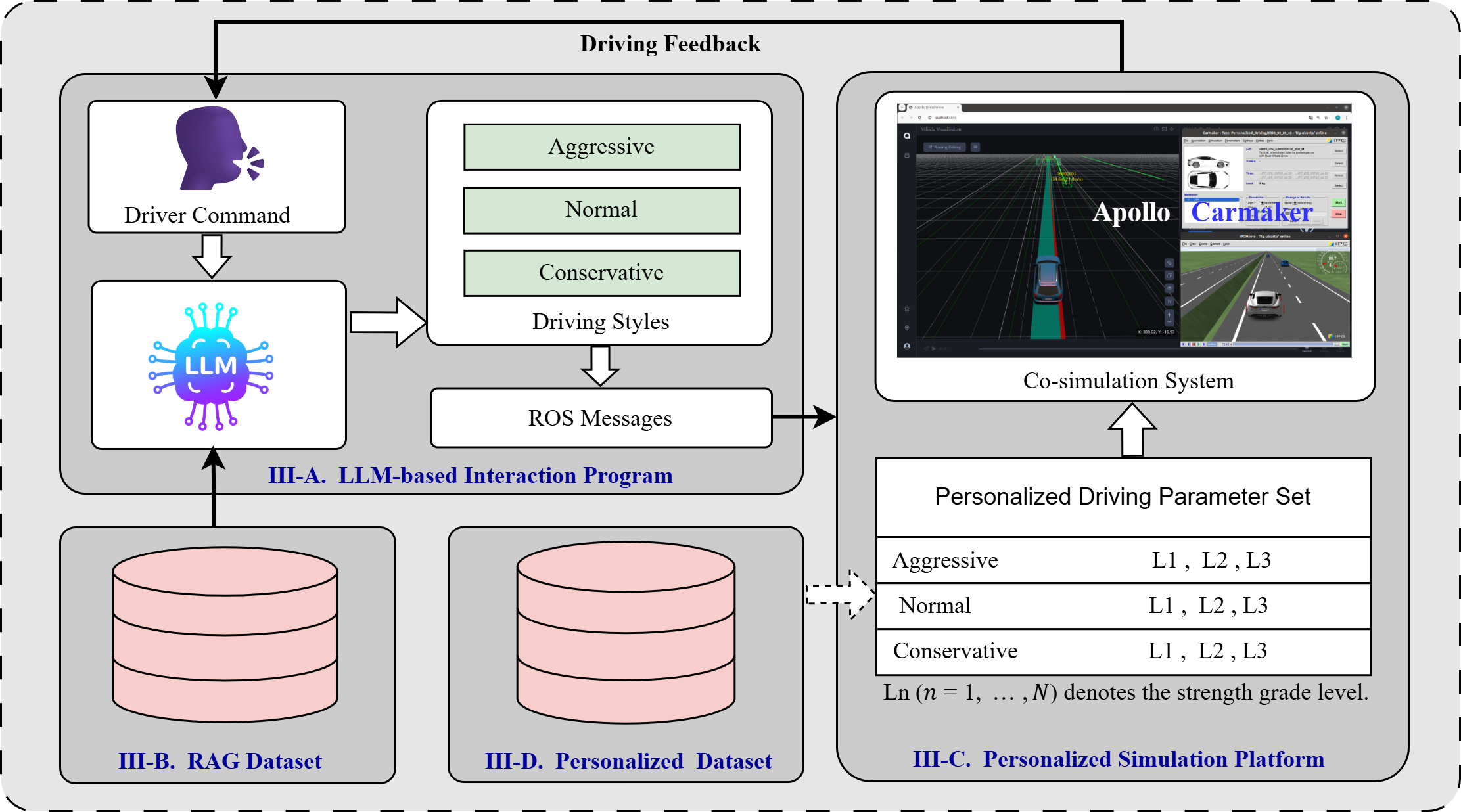}
\caption{Architecture of the proposed personalized driving simulation framework.}
\label{fig_overview}
\end{figure*}

\autoref{fig_overview} illustrates the architecture of the proposed personalized driving simulation framework. The framework enables natural-language interaction between user and automated vehicles, allowing personalized lane-change behaviors to be generated according to user preferences. The LLM-based interaction program, described in Section~\ref{sec:interaction_program}, receives user commands and driving feedback and infers the underlying driving preferences. To improve the interpretation of implicit commands, a retrieval dataset is constructed as described in Section~\ref{sec:rag_dataset}, providing reference examples for LLM inference. The inferred driving styles are then converted into Robot Operating System (ROS) messages and transmitted to the personalized simulation platform described in Section~\ref{sec:simulation_platform}, where the corresponding personalized driving parameter set is selected and executed. To construct this parameter set, a personalized simulation dataset is generated by varying lane-change planning parameters and recording the resulting vehicle behaviors, as described in Section~\ref{sec:personalized_simulation_dataset}. This dataset contains samples generated by varying lane-change planning parameters and recording the corresponding vehicle behaviors.

\subsection{LLM-based Interaction Program}\label{sec:interaction_program}

\begin{figure}[ht]
\centering
\includegraphics[width=.8\columnwidth]{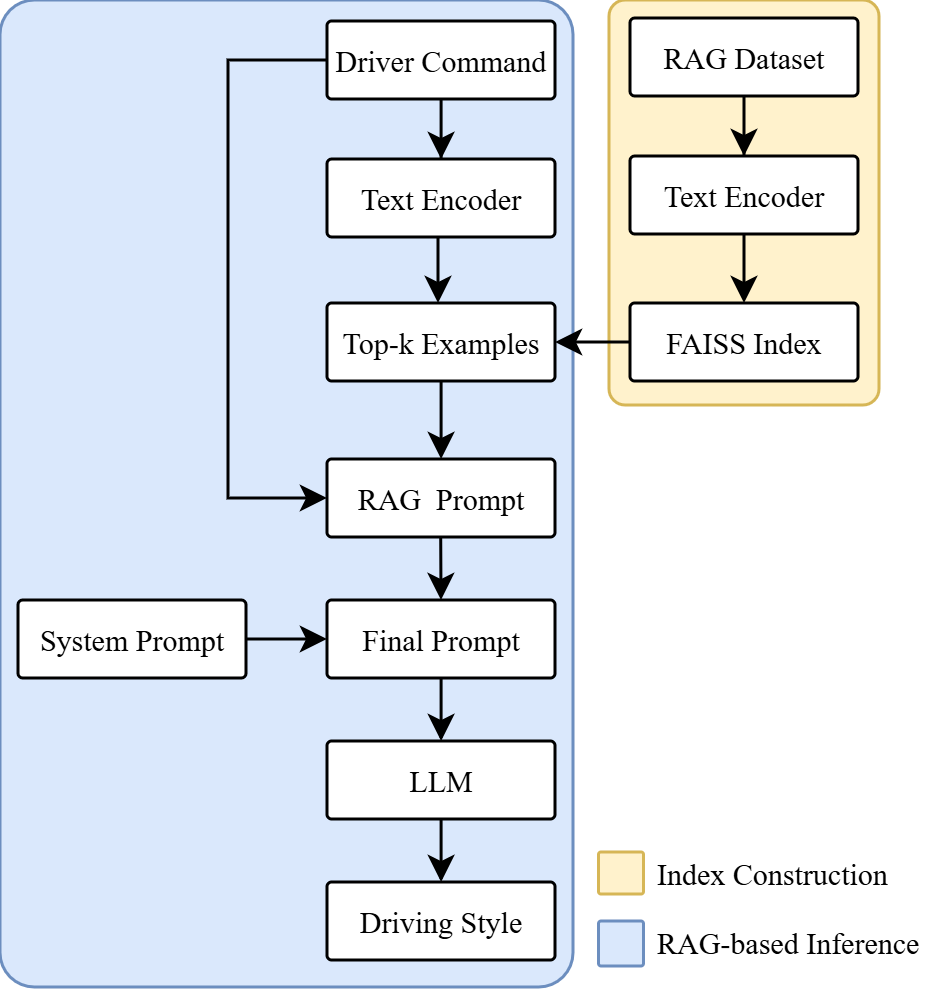}
\caption{Workflow of LLM-based interaction program for driving style classification.}
\label{fig:rag_process}
\end{figure}

\autoref{fig:rag_process} presents the workflow of the proposed LLM-based interaction program, which consists of two main stages: index construction and RAG-based inference. In the index construction stage, the RAG dataset, which contains natural-language preference examples, is encoded into dense embeddings using the \textit{all-MiniLM-L6-v2}~\cite{hf_all_minilm_l6_v2} text encoder, and the embeddings are stored in a Facebook AI Similarity Search (FAISS) index~\cite{douze2024faiss, johnson2019billion}. During RAG-based inference, the user command is encoded by the same text encoder and used as a query to retrieve the top-3 examples from the FAISS index. The retrieved examples are then incorporated into the RAG prompt, which is combined with the system prompt to form the final prompt for LLM-based driving style classification.

\subsection{RAG Dataset}
\label{sec:rag_dataset}
\begin{figure*}[ht]
\centering
\includegraphics[width=7.0in]{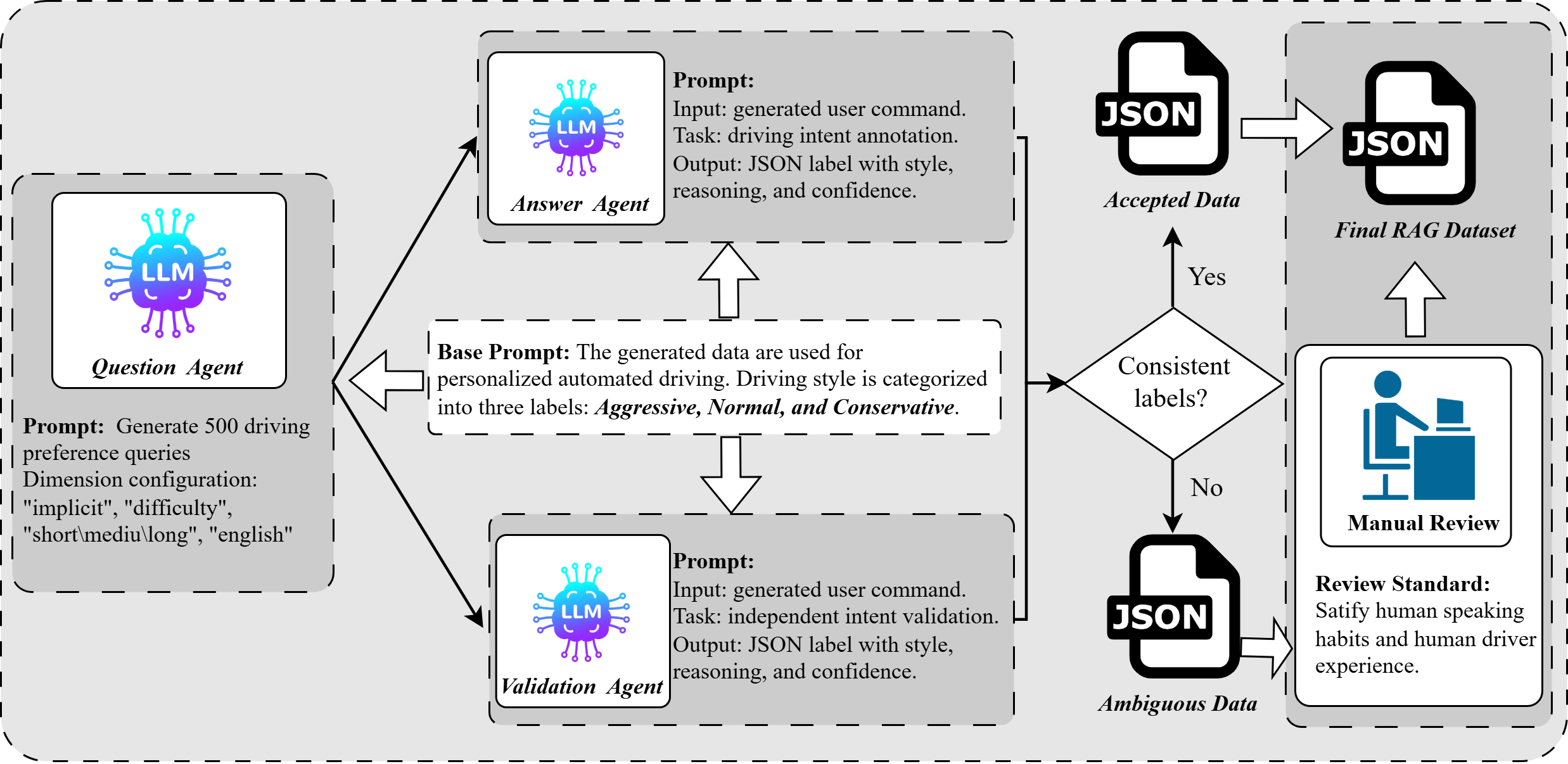}
\caption{Multi-agent data generation framework for constructing the RAG dataset used in driving style classification.}
\label{fig_RAG_Dataset}
\end{figure*}

To support LLM-based driving style classification, a RAG dataset covering driver commands is required. Because manually constructing such data is time-consuming, a multi-agent LLM-based data generation framework is designed, as shown in~\autoref{fig_RAG_Dataset}. The framework follows a "generation-labeling-validation" workflow. First, a question agent generates candidate driving commands according to predefined driving-style targets and command configurations, including explicitness, difficulty, and length. An answer agent then assigns a driving-style label to each command, and a validation agent independently checks the consistency of the generated sample. Specifically, the question agent uses eight different LLMs, as listed in~\autoref{tab:llm_agents}, to improve command diversity. Samples with consistent target labels, answer-agent labels, and validation-agent labels are directly accepted, whereas the remaining samples are assigned to a ambiguous set for manual review.

\begin{table}[!t]
\caption{LLMs Used by Different Agents}
\label{tab:llm_agents}
\centering
\footnotesize
\setlength{\heavyrulewidth}{1.2pt}
\setlength{\lightrulewidth}{0.5pt}
\renewcommand{\arraystretch}{1.2}
\setlength{\tabcolsep}{10pt}
\begin{tabular}{c l}
\toprule
\textbf{Agent} & \hspace{3.0em}\textbf{Models} \\
\midrule
Question Agent 
& \makecell[l]{GPT-4.1; DeepSeek-Chat;\\
Claude-Haiku-4.5;\\
Gemini-2.5-Flash;\\
Grok-4-Fast; Qwen3.5-Plus;\\
Kimi-K2.5; MiniMax-M2.1} \\
\midrule
Answer Agent 
& GPT-4.1 \\
\midrule
Validation Agent 
& DeepSeek-Chat \\
\bottomrule
\end{tabular}
\end{table}

\subsection{Personalized Simulation Platform}
\label{sec:simulation_platform}
The personalized simulation platform consists of two main parts: the co-simulation system and the personalized driving parameter set. After receiving the inferred driving style through ROS messages, the platform selects the corresponding parameter configuration and updates the lane-change planning parameters in the co-simulation system, thereby generating personalized lane-change behaviors.

\subsubsection{Co-simulation System}
A co-simulation system integrating Apollo and CarMaker~\cite{ipg_carmaker} was developed in previous works~\cite{11423564,10588843}. Apollo provides the automated driving stack, whereas CarMaker provides the vehicle dynamics model and simulation environment. The two systems communicate through ROS to exchange vehicle states, planning information, and control commands. In this study, this system serves as the execution environment for personalized lane-change behavior generation, where the selected parameter configuration is applied to the lane-change planning module.

\subsubsection{Tunable Planning Parameters for Personalized Driving}
The personalization capability is achieved by adjusting selected planning parameters in Apollo's longitudinal and lateral planning modules. This subsection identifies the tunable parameters used for personalized lane-change behavior generation. 

\paragraph{Longitudinal Planning Parameters}
Personalized longitudinal behavior during lane-change scenarios is achieved by adjusting the speed planning parameters in Apollo's final speed optimization stage. In this study, two types of longitudinal parameters are considered: objective-function weights and acceleration bounds. The objective-function weights regulate the trade-off among reference tracking, smoothness, and responsiveness, while the acceleration bounds define the allowable acceleration and deceleration range for different driving styles. The complete theoretical analysis process is detailed in Appendix~\ref{app:personalized_longitudinal}.

According to Apollo's planning pipeline, the piecewise-jerk speed optimizer determines the longitudinal speed trajectory by solving
\begin{equation}
\min_{\mathbf{x}} J(\mathbf{x})
=
w_s J_s(\mathbf{x}) + w_v J_v(\mathbf{x}) + w_a J_a(\mathbf{x}) + w_j J_j(\mathbf{x}),
\end{equation}
where $\mathbf{x}=\{s_i,v_i,a_i\}_{i=0}^{N-1}$ denotes the discretized longitudinal trajectory at time step $i$, and $J(\mathbf{x})$ is the total cost of a candidate trajectory. The terms $J_s$, $J_v$, $J_a$, and $J_j$ represent the accumulated costs of position reference tracking, velocity reference tracking, acceleration, and jerk, respectively. The corresponding longitudinal weighting parameters are defined as
\begin{equation}
\mathbf{w}^{\mathrm{lon}} = \{w_s,\, w_v,\, w_a,\, w_j\}.
\end{equation}

In the piecewise-jerk speed optimizer, each weight determines the penalty strength of its corresponding trajectory term. Larger weights impose stronger suppression on the related deviation or motion component. Together with the acceleration bounds, these parameters regulate longitudinal tracking, smoothness, and responsiveness, as summarized in~\autoref{tab_lon_parameters}.

\begin{table*}[!t]
\caption{Personalization Parameters in Longitudinal Speed Planning}
\label{tab_lon_parameters}
\centering
\footnotesize
\setlength{\heavyrulewidth}{1.4pt}
\setlength{\lightrulewidth}{0.5pt}
\renewcommand{\arraystretch}{1.18}
\setlength{\tabcolsep}{5pt}
\begin{tabular*}{0.92\textwidth}{
@{\extracolsep{\fill}}
>{\centering\arraybackslash}m{1.2cm}
>{\centering\arraybackslash}m{3.4cm}
>{\raggedright\arraybackslash}m{9.6cm}
}
\toprule
\textbf{Symbol} & \textbf{Parameter} &
\centering\arraybackslash\textbf{Meaning and Behavioral Effect} \\
\midrule
$\bm{w_s}$ & Position tracking weight
& Tracks the upstream reference position; larger $w_s$ encourages closer adherence to the coarse speed profile. \\
$\bm{w_v}$ & Velocity tracking weight
& Tracks the desired speed; larger $w_v$ encourages stronger cruising-speed tracking. \\
$\bm{w_a}$ & Acceleration weight
& Penalizes longitudinal acceleration; larger $w_a$ suppresses aggressive acceleration and deceleration. \\
$\bm{w_j}$ & Jerk weight
& Penalizes acceleration variation; larger $w_j$ improves smoothness but may reduce responsiveness. \\
\bottomrule
\end{tabular*}
\end{table*}

In addition to the objective-function weights, the acceleration bounds are used to limit the feasible longitudinal motion. For different driving styles, the acceleration constraint is formulated as
\begin{equation}
a_{\min}^{k} \leq a_i \leq a_{\max}^{k},
\quad k \in \{\mathrm{agg}, \mathrm{nor}, \mathrm{con}\},
\end{equation}
where $a_i$ is the longitudinal acceleration, and $k$ denotes the driving style, including aggressive, normal, and conservative styles. The acceleration bounds for different driving styles are predefined according to driving-style-related acceleration characteristics reported in previous studies~\cite{wang2022research}.

\begin{table*}[ht]
\caption{Tunable Lateral Planning Parameters}
\label{tab_lat_parameters}
\centering
\footnotesize
\setlength{\heavyrulewidth}{1.4pt}
\setlength{\lightrulewidth}{0.5pt}
\renewcommand{\arraystretch}{1.18}
\setlength{\tabcolsep}{4pt}
\begin{tabular*}{0.92\textwidth}{
@{\extracolsep{\fill}}
>{\centering\arraybackslash}m{1.1cm}
>{\centering\arraybackslash}m{2.8cm}
>{\centering\arraybackslash}m{2.4cm}
>{\raggedright\arraybackslash}m{7.4cm}
}
\toprule
\textbf{Symbol} & \textbf{Parameter} & \textbf{Category} &
\centering\arraybackslash\textbf{Meaning and Qualitative Effect} \\
\midrule
$\bm{w_l}$ & Lateral offset weight & Objective function
& Penalizes lateral offset; larger $w_l$ discourages large lateral deviations. \\

$\bm{w_{dl}}$ & Slope weight & Objective function
& Penalizes lateral slope; larger $w_{dl}$ suppresses steep lateral transitions. \\

$\bm{w_{ddl}}$ & Curvature weight & Objective function
& Penalizes second-order lateral variation; larger $w_{ddl}$ suppresses second-order lateral variation. \\

$\bm{w_{dddl}}$ & Jerk weight & Objective function
& Penalizes third-order lateral variation; larger $w_{dddl}$ improves smoothness but may reduce responsiveness. \\

$\bm{\delta_{\mathrm{adc}}}$ & Vehicle buffer margin & Geometric constraint
& Expands the initial feasible lateral region; larger $\delta_{\mathrm{adc}}$ improves feasibility but may allow boundary-near paths. \\

$\bm{dl_{\max}}$ & Lateral slope bound & Derivative constraint
& Bounds the lateral derivative; larger $dl_{\max}$ allows steeper and faster lateral transitions. \\
\bottomrule
\end{tabular*}
\end{table*}

\paragraph{Lateral Planning Parameters}
Personalized lateral behavior during lane-change scenarios is achieved by adjusting the path planning parameters in Apollo's piecewise-jerk path optimizer. In this study, three types of lateral parameters are considered: objective-function weights, geometric constraints, and derivative bounds. The objective-function weights regulate the trade-off among lateral offset, lateral slope, curvature-related motion, and high-order smoothness. The geometric constraints define the feasible lateral corridor, while the derivative bounds limit the derivative bounds constrain the lateral slope and curvature-related variation. The complete theoretical analysis process is detailed in Appendix~\ref{app:personalized_lateral}.

According to Apollo's planning pipeline, the piecewise-jerk path optimizer determines the lateral path by solving
\begin{equation}
\begin{aligned}
\min_{\mathbf{y}} J(\mathbf{y})
= {} & w_l J_l(\mathbf{y})
+ w_{dl} J_{dl}(\mathbf{y}) \\
& {} + w_{ddl} J_{ddl}(\mathbf{y})
+ w_{dddl} J_{dddl}(\mathbf{y}),
\end{aligned}
\end{equation}
where $\mathbf{y}=\{l_i,dl_i,ddl_i\}_{i=0}^{N-1}$ denotes the discretized lateral state sequence in the Frenet frame~\cite{werling2010optimal}, and $J(\mathbf{y})$ is the total cost of a candidate lateral path. Here, $l_i$, $dl_i$, and $ddl_i$ represent the lateral offset, first-order lateral derivative, and second-order lateral derivative, respectively. The terms $J_l$, $J_{dl}$, $J_{ddl}$, and $J_{dddl}$ represent the accumulated costs of lateral offset, lateral slope, second-order lateral variation, and high-order lateral smoothness, respectively. The corresponding lateral weighting parameters are defined as
\begin{equation}
\mathbf{w}^{\mathrm{lat}} = \{w_l,\, w_{dl},\, w_{ddl},\, w_{dddl}\}.
\end{equation}

In addition to the objective-function weights, the feasible lateral region is constrained by the lane boundaries, obstacle constraints, and vehicle safety margin. The geometric constraint is expressed as
\begin{equation}
l_{\min}(\xi_i;\delta_{\mathrm{adc}}) \leq l_i \leq l_{\max}(\xi_i;\delta_{\mathrm{adc}}),
\end{equation}
where $\xi_i$ is the spatial sampling point along the reference line, $l_i$ is the lateral offset at $\xi_i$, and $\delta_{\mathrm{adc}}$ denotes the vehicle buffer margin used to adjust the feasible lateral corridor.

The lateral transition is further limited by the lateral slope bound
\begin{equation}
|dl_i| \leq dl_{\max}.
\end{equation}
where $dl_i$ denotes the first-order lateral derivative with respect to the reference-line arc length, and $\bm{dl_{\max}}$ defines the maximum allowable lateral slope.

\begin{table*}[!b]
\caption{Representative Parameter Mapping for Personalized Driving Styles}
\label{tab:style_param_template}
\centering
\footnotesize
\setlength{\heavyrulewidth}{1.2pt}
\setlength{\lightrulewidth}{0.5pt}
\renewcommand{\arraystretch}{1.2}
\setlength{\tabcolsep}{4pt}

\begin{tabular*}{\textwidth}{@{\extracolsep{\fill}} c c cccccc cccccc}
\toprule
\multirow{2}{*}{\textbf{Style}} 
& \multirow{2}{*}{\textbf{Level}}
& \multicolumn{6}{c}{\textbf{Longitudinal Parameters}}
& \multicolumn{6}{c}{\textbf{Lateral Parameters}} \\
\cmidrule(lr){3-8} \cmidrule(lr){9-14}
& 
& $w_s$ & $w_v$ & $w_a$ & $w_j$ & $a_{\min}$ & $a_{\max}$
& $w_l$ & $w_{dl}$ & $w_{ddl}$ & $w_{dddl}$ & $\delta_{\mathrm{adc}}$ & $dl_{\max}$ \\
\midrule

\multirow{3}{*}{Aggressive}
& L1 & 0.6 & 30 & 0.3 & 0.2 & -3.0 & 3.0 & 3.0 & 5.0 & 200 & 5000 & 0.5 & 3.0 \\
& L2 & 0.6 & 30 & 0.3 & 0.2 & -3.0 & 3.0 & 2.0 & 4.0 & 100 & 3000 & 0.3 & 3.0 \\
& L3 & 0.8 & 25 & 0.5 & 0.2 & -3.0 & 3.0 & 1.0 & 4.0 & 300 & 8000 & 0.5 & 2.0 \\
\midrule

\multirow{3}{*}{Normal}
& L1 & 0.8 & 15 & 0.7 & 1.5 & -2.75 & 2.75 & 0.7 & 10 & 700 & 12000 & 0.7 & 1.2 \\
& L2 & 0.8 & 15 & 0.7 & 1.5 & -2.75 & 2.75 & 0.7 & 10 & 500 & 20000 & 0.7 & 1.2 \\
& L3 & 0.8 & 15 & 0.7 & 1.5 & -2.75 & 2.75 & 0.7 & 12 & 700 & 20000 & 0.7 & 1.6 \\
\midrule

\multirow{3}{*}{Conservative}
& L1 & 1.0 & 10 & 1.0 & 3.0 & -2.5 & 2.5 & 0.4 & 20 & 1200 & 20000 & 1.0 & 0.5 \\
& L2 & 1.0 & 10 & 1.5 & 3.0 & -2.5 & 2.5 & 0.2 & 20 & 1200 & 50000 & 1.0 & 0.4 \\
& L3 & 1.0 & 10 & 1.5 & 3.0 & -2.5 & 2.5 & 0.4 & 20 & 1200 & 60000 & 1.0 & 1.0 \\
\bottomrule
\end{tabular*}
\end{table*}

In the piecewise-jerk path optimizer, each weight determines the penalty strength of its corresponding lateral trajectory term. Together with the geometric constraints and derivative bounds, these parameters regulate the lateral aggressiveness, smoothness, and feasibility of lane-change trajectories, as summarized in~\autoref{tab_lat_parameters}.

\paragraph{Personalized Parameter Mapping}
Based on these selected longitudinal and lateral planning parameters, a personalized driving parameter mapping is constructed to associate each driving style with multiple executable parameter configurations. As summarized in \autoref{tab:style_param_template}, each style is represented by three intensity levels, denoted as \(L_1\), \(L_2\), and \(L_3\). These levels are obtained through the intra-class ranking procedure described in Section~\ref{sec:personalized_simulation_dataset}, where a higher level indicates a stronger expression of the corresponding driving style.

\subsection{Personalized Dataset}
\label{sec:personalized_simulation_dataset}
This subsection presents the construction process of the personalized dataset. A highway lane-change scenario is first defined in the co-simulation system. Then, selected Apollo planning parameters are varied within feasible ranges to generate offline simulation samples and record the corresponding lane-change behaviors. Based on the valid samples, clustering is used to identify aggressive, normal, and conservative driving styles, and intra-class ranking is further applied to select representative parameter sets with different intensity levels.

\subsubsection{Lane-Change Scenario}
\begin{figure}[ht]
\centering
\includegraphics[width=\columnwidth]{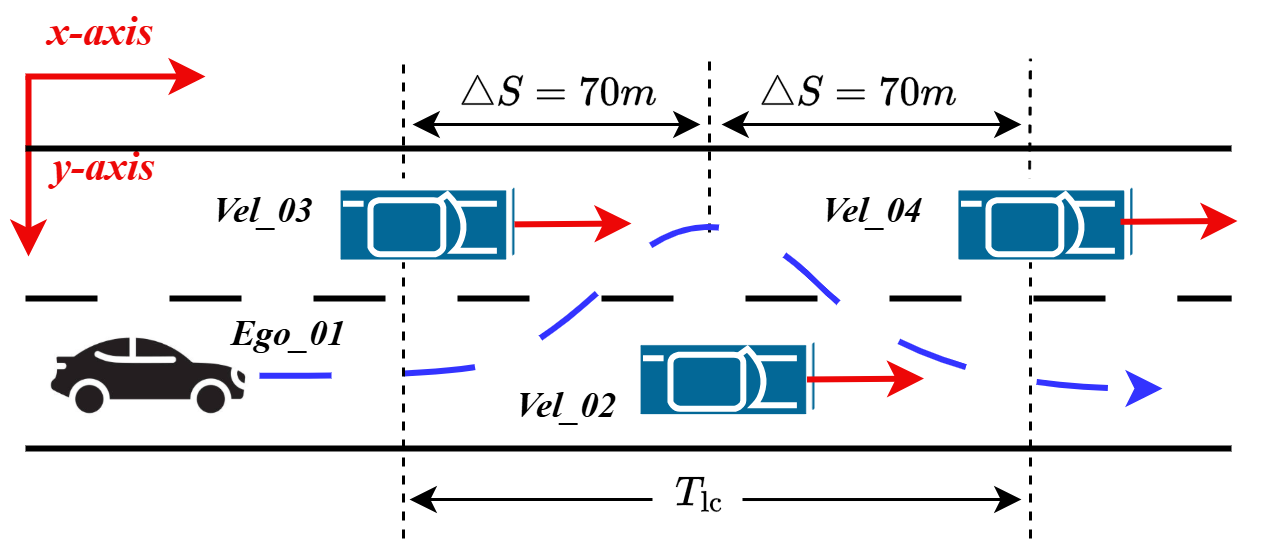}
\caption{A typical lane change scenario.}
\label{fig_2}
\end{figure}
As shown in~\autoref{fig_2}, the scenario includes ego vehicle and three surrounding vehicles with predefined  parameters. The ego vehicle is initialized at 126~km/h, while all surrounding vehicles travel at 80~km/h. The initial longitudinal spacing is determined according to time-to-collision (TTC) criterion. A target TTC of 5.5 s is adopted, which is slightly larger than the 4.5–5 s range suggested for motorway collision-avoidance warning strategies [41]. The spacing is calculated as
\begin{equation}
d = v_{\mathrm{rel}} \cdot TTC.
\end{equation}
where $v_{\mathrm{rel}}$ denotes the relative speed between the ego vehicle and the traffic vehicle.

\subsubsection{Offline Simulation}
\autoref{tab_unified_param} lists the input parameter ranges and the corresponding behavioral outputs recorded from each simulation. Each selected input parameter is discretized into several representative values with equal spacing for offline dataset generation.

\begin{table}[!t]
\caption{Personalized Parameter Space and Behavioral Effects}
\label{tab_unified_param}
\centering
\footnotesize
\setlength{\heavyrulewidth}{1.2pt}
\setlength{\lightrulewidth}{0.5pt}
\renewcommand{\arraystretch}{1.2}
\setlength{\tabcolsep}{4pt}
\begin{tabular}{
c
|>{\centering\arraybackslash}m{2.0cm}
>{\centering\arraybackslash}m{2.0cm}|
>{\centering\arraybackslash}m{2.8cm}
}
\toprule
\multicolumn{1}{c}{\textbf{Dim.}} &
\multicolumn{1}{c}{\textbf{Parameters}} &
\multicolumn{1}{c}{\textbf{Range}} &
\multicolumn{1}{c}{\textbf{Output}} \\
\midrule
Long.
&
\begin{tabular}{c}
$w_s$ \\
$w_v$ \\
$w_a$ \\
$w_j$ \\
$a_{\min}$ \\
$a_{\max}$
\end{tabular}
&
\begin{tabular}{c}
$[0.6,\,1.2]$ \\
$[10,\,30]$ \\
$[0.3,\,1.5]$ \\
$[0.2,\,3.0]$ \\
$[-3.0,\,-2.5]$ \\
$[2.5,\,3.0]$
\end{tabular}
&
\begin{tabular}{c}
$T_{\mathrm{lc}}$ (s) \\
$v_{\mathrm{ego}}$ (m/s) \\
$a_{\mathrm{x}}$ (m/s$^2$) \\
$X_j$ (m),\; $j=1,2,3,4$
\end{tabular}
\\
\midrule
Lat.
&
\begin{tabular}{c}
$w_l$ \\
$w_{dl}$ \\
$w_{ddl}$ \\
$w_{dddl}$ \\
$\delta_{\mathrm{adc}}$ \\
$dl_{\max}$
\end{tabular}
&
\begin{tabular}{c}
$[0.1,\,4.0]$ \\
$[5,\,20]$ \\
$[100,\,1200]$ \\
$[3000,\,50000]$ \\
$[0.3,\,1.0]$ \\
$[0.4,\,4.0]$
\end{tabular}
&
\begin{tabular}{c}
$\delta_{\mathrm{sw}}$ (rad) \\
$r_{\mathrm{ego}}$ (rad/s) \\
$a_{\mathrm{y}}$ (m/s$^2$) \\
$Y_j$ (m),\; $j=1,2,3,4$
\end{tabular}
\\
\bottomrule
\end{tabular}
\vspace{1mm}
\begin{minipage}{0.96\columnwidth}
\scriptsize
\end{minipage}
\end{table}

The ranges of the selected longitudinal and lateral planning parameters are determined from Apollo's default configuration and simulation-based boundary exploration. Starting from the default values, parameter-sweeping simulations are performed to identify settings that can maintain valid lane-change behavior. Settings causing planning failure, unsuccessful lane changes, speed-tracking failure, repeated trajectory corrections, or road-boundary violations are excluded. The remaining values are used as engineering ranges for offline dataset generation.

The output variables are selected to capture the personalized lane-change behavior produced by different parameter settings. The lane-change time $T_{\mathrm{lc}}$ is used to describe the temporal efficiency of the maneuver, as illustrated in ~\autoref{fig_2}. The ego velocity $v_{\mathrm{ego}}$ and longitudinal acceleration $a_x$ characterize the longitudinal speed-tracking and acceleration behavior during the lane-change process. For lateral behavior, Previous research collected vehicle-state signals such as lateral acceleration $a_y$, steering wheel angle $\delta_{\mathrm{sw}}$, and yaw rate $r_{\mathrm{ego}}$, and used their characteristic values for driver-behavior classification into cautious, normal, and aggressive groups~\cite{8449120}. In addition, the absolute positions of the ego vehicle and surrounding vehicles, denoted as $X_j$ and $Y_j$, are recorded to characterize the spatial evolution of the lane-change process and the surrounding traffic context.

\subsubsection{Clustering}

\begin{table}[!b]
\caption{Dataset Generation and Clustering Statistics}
\label{tab:dataset_summary}
\centering
\footnotesize
\setlength{\heavyrulewidth}{1.2pt}
\setlength{\lightrulewidth}{0.5pt}
\renewcommand{\arraystretch}{1.2}
\setlength{\tabcolsep}{6pt}   
\begin{tabular}{
>{\centering\arraybackslash}p{0.32\columnwidth}
>{\centering\arraybackslash}p{0.43\columnwidth}
c
}
\toprule
\textbf{Group} & \textbf{Category} & \textbf{Number} \\
\midrule
\multirow{3}{*}{Dataset Generation}
 & Total samples & 465 \\
 & Valid samples & 425 \\
 & Invalid samples & 40 \\
\midrule
\multirow{3}{*}{\makecell{\textbf{Clustering Results}\\\footnotesize (K-means, $K=3$)}}
 & Aggressive & 97 \\
 & Normal & 175 \\
 & Conservative & 153 \\
\bottomrule
\end{tabular}
\end{table}

\begin{table}[!t]
\caption{Clustering Features for Driving Style Identification}
\label{tab:clustering_features}
\centering
\footnotesize
\setlength{\heavyrulewidth}{1.2pt}
\setlength{\lightrulewidth}{0.5pt}
\renewcommand{\arraystretch}{1.2}
\setlength{\tabcolsep}{7pt}

\begin{tabular}{
>{\centering\arraybackslash}m{2.4cm}
>{\centering\arraybackslash}m{2.2cm}
>{\centering\arraybackslash}m{1.3cm}
}
\toprule
\textbf{Category} & \textbf{Feature} & \textbf{Weight} \\
\midrule
Lane change time 
& $T_{\mathrm{lc}}$ 
& 25.40\% \\
\midrule
\multirow{2}{*}{Longitudinal acceleration}
& $\max |a_x|$ 
& 3.67\% \\
& $E_{a_x}$ 
& 3.67\% \\
\midrule
\multirow{2}{*}{Lateral acceleration}
& $\max |a_y|$ 
& 15.04\% \\
& $E_{a_y}$ 
& 16.60\% \\
\midrule
\multirow{2}{*}{Steering wheel angle}
& $\max |\delta_{\mathrm{sw}}|$ 
& 6.08\% \\
& $E_{\delta_{\mathrm{sw}}}$ 
& 11.55\% \\
\midrule
\multirow{2}{*}{Yaw rate}
& $\max |r_{\mathrm{ego}}|$ 
& 6.81\% \\
& $E_{r_{\mathrm{ego}}}$ 
& 11.18\% \\
\bottomrule
\end{tabular}
\end{table}

A detailed statistical summary of the dataset generation and clustering results is provided in~\autoref{tab:dataset_summary}. 488 simulation samples were generated, among which 425 were valid and 40 were identified as invalid samples. K-means clustering was then applied to the valid samples to identify three representative driving styles. K-means partitions samples into $K$ clusters by minimizing the within-cluster distance between each sample and its assigned cluster centroid~\cite{mcqueen1967some}. In this study, $K$ is set to 3 to obtain aggressive, normal, and conservative driving styles.

For driving-style identification, the clustering feature vector is constructed from the longitudinal and lateral dynamic characteristics of each lane-change maneuver. The selected features are summarized in~\autoref{tab:clustering_features}, where the accumulated squared response $E_q$ is used to describe the overall intensity of each dynamic variable. Here, $q$ denotes a generic signal selected from $a_x$, $a_y$, $\delta_{\mathrm{sw}}$, and $r_{\mathrm{ego}}$.

For a dynamic variable $q(t)$, the accumulated squared response is defined as
\begin{equation}
E_q = \int_{t_{\mathrm{start}}}^{t_{\mathrm{end}}} q^2(t)\,dt,
\end{equation}
where $t_{\mathrm{start}}$ and $t_{\mathrm{end}}$ denote the start and end times of the lane-change maneuve. This term represents the overall intensity of the corresponding motion response during the maneuver. A larger value indicates a stronger or more persistent dynamic response.

\begin{figure}[ht]
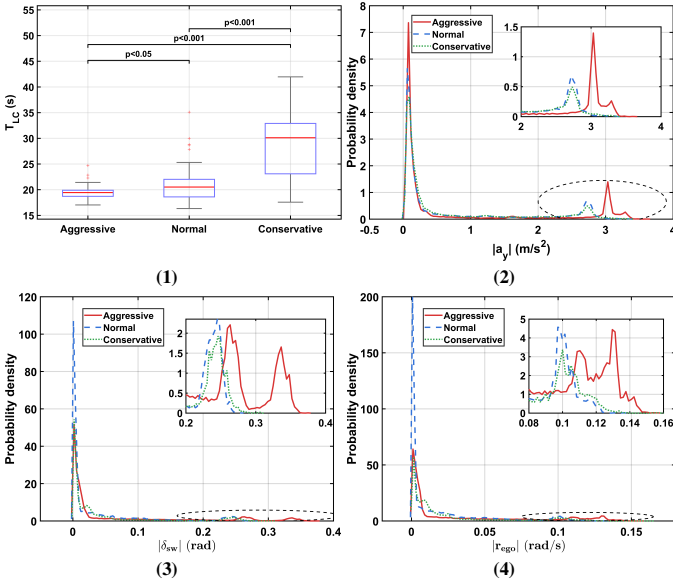

\centering

% ---------- top row ----------
\begin{minipage}{\columnwidth}
    \centering
    \includegraphics[width=\linewidth]{Pictures/lanechange_statistics_top_1x2.png}

    \vspace{-1mm}
    \makebox[\linewidth]{%
        \begin{minipage}{0.49\linewidth}
            \centering
            {\scriptsize \textbf{(1)}}
        \end{minipage}
        \begin{minipage}{0.49\linewidth}
            \centering
            {\scriptsize \textbf{(2)}}
        \end{minipage}
    }
\end{minipage}

\vspace{1mm}

% ---------- bottom row ----------
\begin{minipage}{\columnwidth}
    \centering
    \includegraphics[width=\linewidth]{Pictures/lanechange_statistics_bottom_1x2.png}

    \vspace{-1mm}
    \makebox[\linewidth]{%
        \begin{minipage}{0.49\linewidth}
            \centering
            {\scriptsize \textbf{(3)}}
        \end{minipage}
        \begin{minipage}{0.49\linewidth}
            \centering
            {\scriptsize \textbf{(4)}}
        \end{minipage}
    }
\end{minipage}

\vspace{-1mm}
\caption{Statistical distributions of lane-change time and lateral dynamic responses across driving styles.}
\label{fig_pdfkmeanresul}
\end{figure}

\autoref{fig_pdfkmeanresul} provides a visual illustration of the sample distributions for the three driving styles obtained via clustering. Subfigure~(1) shows that the lane-change time increases from aggressive to conservative driving styles, with statistically significant differences between the groups based on Student's $t$-test ($p<0.05$ and $p<0.001$). Subfigures~(2)--(4) show the probability distributions of lateral acceleration, steering wheel angle, and yaw rate during lane changes. The aggressive style generally presents larger values and broader distributions, indicating stronger lateral dynamic responses. In contrast, the conservative style shows more concentrated distributions with lower magnitudes, reflecting smoother lane-change behavior. The normal style lies between these two extremes. These results indicate that the constructed dataset can effectively distinguish different driving styles through both lane-change duration and lateral dynamic characteristics.

\subsubsection{Style-Intensity Ranking}Based on the personalized dataset obtained from clustering, the diversity of input parameter combinations and the varying influence of different parameters on personalized lane-change behavior may lead to insufficiently distinguishable differences among samples. To further evaluate the proposed framework’s ability to distinguish not only different driving styles but also different style intensities within the same driving style, three representative control parameter sets with different intensity levels are selected from each driving style.

To rank the samples within each driving-style category, the behavioral features are first normalized using z-score standardization to eliminate the influence of different units and numerical scales
\begin{equation}
z_{ij}=\frac{x_{ij}-\mu_j}{\sigma_j},
\end{equation}
where $x_{ij}$ denotes the original value of the $j$-th behavioral feature of the $i$-th sample, and $\mu_j$ and $\sigma_j$ are the mean and standard deviation of the $j$-th feature over all valid samples, respectively. After standardization, the $i$-th sample is represented by the standardized behavioral feature vector
\begin{equation}
\mathbf{z}_i=[z_{i1},z_{i2},\ldots,z_{ip}],
\end{equation}
where $p$ is the number of behavioral features.

Since each standardized feature describes the deviation of a sample from the average value of that feature, the distance from a sample to the origin of the standardized feature space is used to quantify its overall deviation from average lane-change behavior. This distance is defined as the style-intensity score
\begin{equation}
S_i=\|\mathbf{z}_i\|_2
=\sqrt{z_{i1}^2+z_{i2}^2+\cdots+z_{ip}^2}.
\end{equation}
A larger $S_i$ indicates a stronger deviation from average lane-change behavior and is therefore interpreted as a higher style intensity.

Within each driving-style category, samples are ranked according to $S_i$. Three representative samples are then selected to correspond to low, medium, and high intensity levels.

\section{EXPERIMENT}
%See \cite{ref1,ref2,ref3,ref4,ref5} for resources on formatting math into text and additional help in working with \LaTeX .
The experiments are designed to evaluate the effectiveness of the RAG dataset and the lane-change behavior differences under the nine style-parameter configurations defined in~\autoref{tab:style_param_template}. As summarized in~\autoref{tab:experimental_design}, the experiments consist of an LLM-based preference interpretation test and a system-level behavior test. The interpretation test uses explicit, mixed, and implicit command sets, where the mixed set contains 50\% explicit and 50\% implicit commands. The system-level test evaluates the resulting vehicle behaviors under different driving styles and intensity levels. For each test run, the lane-change events along the route are extracted and statistically evaluated using lane-change time, maximum lateral acceleration, and accumulated lateral-acceleration response.

In addition to the quantitative experiments, a complete usage example is provided online~\href{https://youtu.be/te6xv4aDV6I}{[Video]} to demonstrate the workflow from natural-language command input, LLM-based style inference, and parameter selection to co-simulation-based behavior execution.

\begin{table}[ht]
\caption{Experimental Design}
\label{tab:experimental_design}
\centering
\footnotesize
\setlength{\heavyrulewidth}{1.2pt}
\setlength{\lightrulewidth}{0.5pt}
\renewcommand{\arraystretch}{1.15}
\setlength{\tabcolsep}{4pt}

\begin{tabular}{@{}c c l c@{}}
\toprule
\textbf{Part} & \textbf{No.} & \textbf{Test Condition} & \textbf{Evaluation} \\
\midrule

\multirow{3}{*}{\makecell[c]{LLM\\interpretation\\test}}
& 1 & explicit samples & \multirow{3}{*}{Accuracy} \\
& 2 & mixed samples &  \\
& 3 & implicit samples &  \\

\midrule

\multirow{9}{*}{\makecell[c]{System-level\\behavior test}}
& 1 & Aggressive style, L1 & \multirow{9}{*}{\makecell{$\overline{T}_{\mathrm{LC}}$,\\ $\max |a_y|$,\\ $\overline{E}_{a_y}$}} \\
& 2 & Aggressive style, L2 &  \\
& 3 & Aggressive style, L3 &  \\
& 4 & Normal style, L1 &  \\
& 5 & Normal style, L2 &  \\
& 6 & Normal style, L3 &  \\
& 7 & Conservative style, L1 &  \\
& 8 & Conservative style, L2 &  \\
& 9 & Conservative style, L3 &  \\

\bottomrule
\end{tabular}
\end{table}

\subsection{Performance Comparison of Zero-Shot and RAG-Based Preference Interpretation}

Three test settings are designed to evaluate LLM-based driving-style classification under different levels of command explicitness. Each test set contains 80 samples. As summarized in~\autoref{tab:rag_results_combined}, RAG-based inference improves classification accuracy across all three test settings. For the explicit command set, both zero-shot and RAG-based inference achieve high accuracy, indicating that explicit style commands can already be accurately interpreted by most LLMs. For the mixed command set, RAG provides consistent gains, showing its benefit when explicit and implicit expressions appear together. The largest improvement is observed for the implicit command set, where the average gain reaches 10.2 percentage points, indicating that retrieved examples are particularly helpful for interpreting indirect user commands.

\begin{table}[ht]
\caption{Accuracy Comparison on Explicit, Mixed, and Implicit Command Sets}
\label{tab:rag_results_combined}
\centering
\scriptsize
\setlength{\heavyrulewidth}{1.2pt}
\setlength{\lightrulewidth}{0.5pt}
\renewcommand{\arraystretch}{1.12}
\setlength{\tabcolsep}{2pt}
\begin{tabular*}{\columnwidth}{
@{\extracolsep{\fill}}
>{\centering\arraybackslash}p{0.12\columnwidth}
>{\centering\arraybackslash}p{0.34\columnwidth}
>{\centering\arraybackslash}p{0.16\columnwidth}
>{\centering\arraybackslash}p{0.14\columnwidth}
>{\centering\arraybackslash}p{0.12\columnwidth}
}
\toprule
\textbf{Set} & \textbf{Model} & \textbf{Zero-shot} & \textbf{RAG} & \textbf{Gain} \\
\midrule
\multirow{8}{*}{Explicit}
& Gemma-4-31B-IT & 97.5\% & 98.8\% & +1.3 \\
& Qwen-2.5-7B-Instruct & 97.5\% & 98.8\% & +1.3 \\
& Llama-3.1-8B-Instruct & 87.5\% & 95.0\% & +7.5 \\
& Ministral-8B-2512 & 98.8\% & 98.8\% & +0.0 \\
& Gemma-3-4B-IT & 80.0\% & 88.8\% & +8.8 \\
& Qwen3-8B & 97.5\% & 98.8\% & +1.3 \\
& Gemma-3-12B-IT & 96.2\% & 100.0\% & +3.8 \\
\cmidrule(lr){2-5}
& \textbf{Average} & \textbf{93.6\%} & \textbf{97.0\%} & \textbf{+3.4} \\
\midrule
\multirow{8}{*}{Mixed}
& Gemma-4-31B-IT & 93.8\% & 95.0\% & +1.2 \\
& Qwen-2.5-7B-Instruct & 86.2\% & 92.5\% & +6.3 \\
& Llama-3.1-8B-Instruct & 86.2\% & 91.2\% & +5.0 \\
& Ministral-8B-2512 & 86.2\% & 87.5\% & +1.3 \\
& Gemma-3-4B-IT & 81.2\% & 91.2\% & +10.0 \\
& Qwen3-8B & 81.2\% & 87.5\% & +6.3 \\
& Gemma-3-12B-IT & 83.8\% & 96.2\% & +12.4 \\
\cmidrule(lr){2-5}
& \textbf{Average} & \textbf{85.5\%} & \textbf{91.6\%} & \textbf{+6.1} \\
\midrule
\multirow{8}{*}{Implicit}
& Gemma-4-31B-IT & 83.8\% & 86.2\% & +2.4 \\
& Qwen-2.5-7B-Instruct & 71.2\% & 87.5\% & +16.3 \\
& Llama-3.1-8B-Instruct & 71.2\% & 80.0\% & +8.8 \\
& Ministral-8B-2512 & 67.5\% & 77.5\% & +10.0 \\
& Gemma-3-4B-IT & 68.8\% & 78.8\% & +10.0 \\
& Qwen3-8B & 67.5\% & 81.2\% & +13.7 \\
& Gemma-3-12B-IT & 72.5\% & 82.5\% & +10.0 \\
\cmidrule(lr){2-5}
& \textbf{Average} & \textbf{71.8\%} & \textbf{82.0\%} & \textbf{+10.2} \\
\bottomrule
\end{tabular*}
\\[2pt]
\parbox{\columnwidth}{\raggedright\scriptsize Gain denotes the accuracy difference between RAG and zero-shot prompting, measured in percentage points.}
\end{table}

\subsection{System-Level Evaluation of Personalized Lane-Change Behavior}

\begin{figure}[!b]
\centering
\includegraphics[width=\columnwidth]{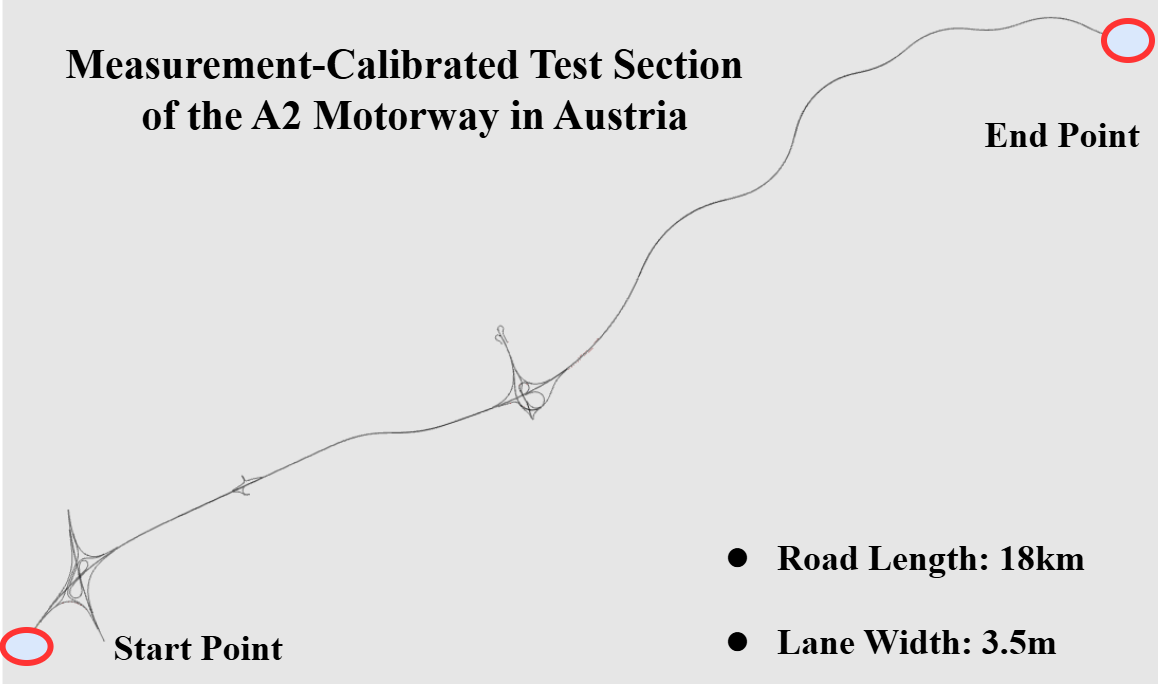}
\caption{Road network used for the experiments.}
\label{fig_A2map}
\end{figure}

The experiments are conducted on a road network reconstructed from a motorway in Austria~\cite{8916839,11251171}, as shown in~\autoref{fig_A2map}. Background traffics are generated using CarMaker's stochastic traffic-flow function, where vehicles are randomly distributed along predefined routes according to a relative density parameter. A total of nine experiments were conducted, where each experiment corresponds to a complete driving process from the starting point to the destination. 

\begin{figure}[!t]
\centering
\includegraphics[width=\columnwidth]{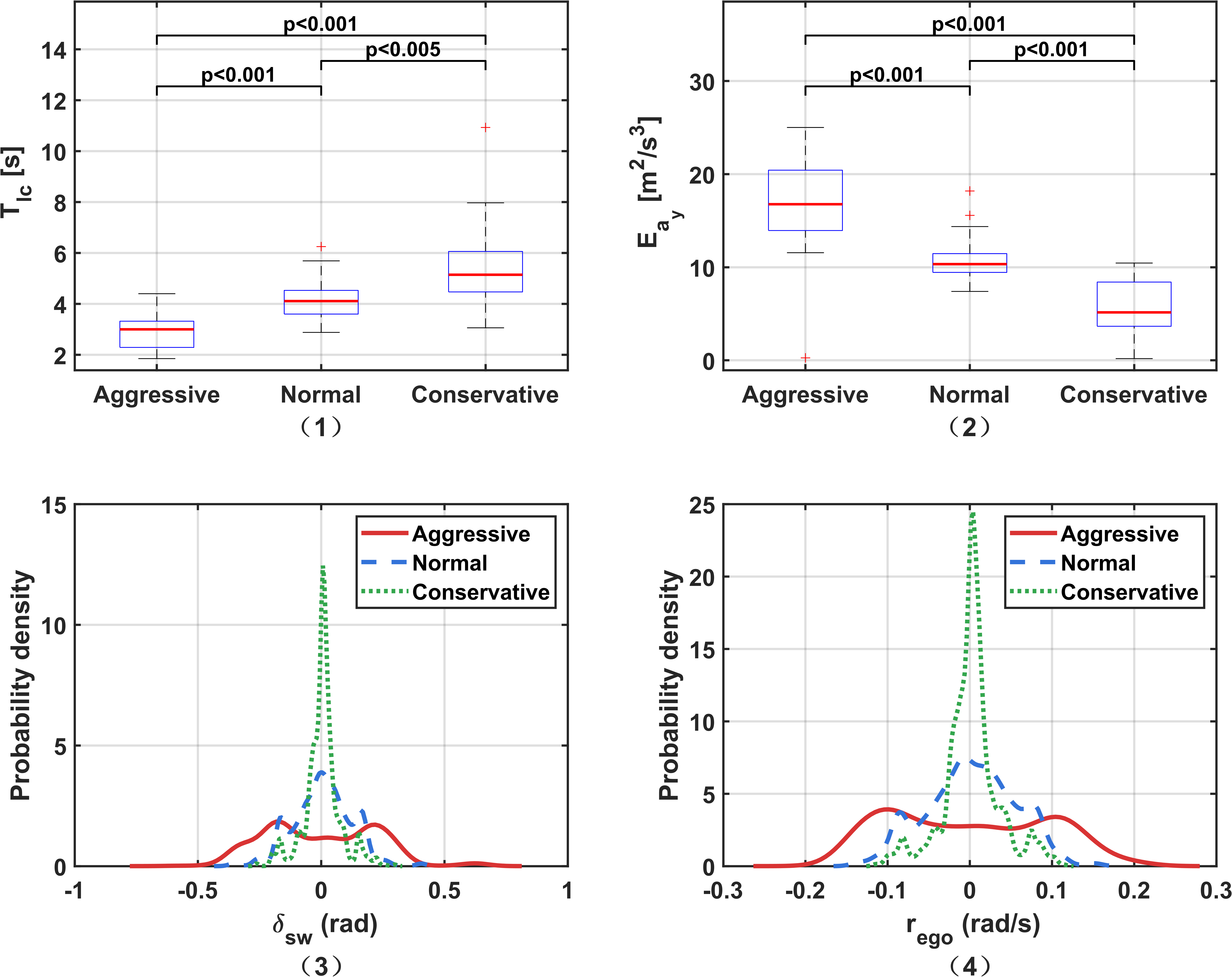}
\caption{Lane-change KPI distributions across driving styles.}
\label{fig_pdffinalresul}
\end{figure}

The results are analyzed from two perspectives. First, \autoref{fig_pdffinalresul} groups the extracted lane-change events by driving style and presents the statistical distributions of aggressive, normal, and conservative behaviors. Second,~\autoref{tab:lc_kpi_summary} further summarizes the results by both driving style and intensity level, enabling a comparison among L1, L2, and L3 within each driving style.

\begin{table}[ht]
\caption{Lane-change KPIs across driving styles on the A2 motorway.
$\uparrow$: higher values indicate a more intensive lane-change dynamic process;
$\downarrow$: lower values indicate a gentler lane-change dynamic process.}
\label{tab:lc_kpi_summary}
\centering
\footnotesize
\setlength{\heavyrulewidth}{1.2pt}
\setlength{\lightrulewidth}{0.5pt}
\renewcommand{\arraystretch}{1.18}

\begin{tabular*}{\columnwidth}{@{\extracolsep{\fill}} l c c c c}
\toprule
\textbf{Style} &
\makecell{\textbf{Strength}\\\textbf{Grade}} &
\multicolumn{3}{c}{\textbf{Lane-change performance}} \\
\cmidrule(lr){3-5}
& &
$\overline{T}_{\mathrm{LC}} \downarrow$ [s] &
$\max |a_y| \uparrow$ &
$\overline{E}_{a_y} \uparrow$ \\
\midrule

\multirow{3}{*}{Aggressive}
& L1 & \textbf{2.4790} & \textbf{3.5362} & \textbf{17.6299} \\
& L2 & 3.3800 & 3.3918 & 16.4673 \\
& L3 & 3.7150 & 3.3688 & 14.1935 \\

\midrule
\multirow{3}{*}{Normal}
& L1 & 3.7675 & 3.1043 & 12.0857 \\
& L2 & 4.2017 & 2.9980 & 10.4531 \\
& L3 & 4.4467 & 2.9901 & 9.2197 \\

\midrule
\multirow{3}{*}{Conservative}
& L1 & 5.1692 & 2.8463 & 7.6423 \\
& L2 & 5.5450 & 2.6409 & 4.1332 \\
& L3 & 6.1725 & 2.0355 & 2.4949 \\

\bottomrule
\end{tabular*}
\end{table}

\autoref{fig_pdffinalresul} shows clear behavioral differences among the three driving styles. The aggressive style generally presents shorter lane-change time and stronger lateral dynamic responses, including larger lateral acceleration intensity, wider steering-wheel-angle distributions, and higher yaw-rate values. In contrast, the conservative style shows longer lane-change time and more concentrated distributions around smaller lateral acceleration, steering-wheel angle, and yaw rate values, indicating smoother and more stable lane-change behavior. The normal style generally lies between these two extremes.~\autoref{tab:lc_kpi_summary} presents the statistical analysis of representative personalized lane-change KPIs. As the style shifts from aggressive to conservative, the lane-change time increases, while both the maximum lateral acceleration and lateral acceleration intensity decrease, indicating smoother maneuvers.

\section{Conclusion}

This study proposed an LLM-enabled personalized lane-change framework for highway automated driving. The framework maps natural-language driver commands to personalized driving styles and executable planning parameters within the Apollo planning pipeline. The results show that different parameter settings generate distinguishable lane-change behaviors in lane-change time, lateral acceleration, steering response, and yaw-rate characteristics. 

To improve the interpretation of implicit preference commands, a RAG dataset was constructed for retrieval-augmented generation. Three test settings were used to compare zero-shot and RAG-based inference across multiple LLMs. The results show that RAG provides consistent improvements, especially for implicit preference commands and relatively smaller models. This indicates that retrieved implicit examples can provide useful semantic guidance for mapping indirect natural-language expressions to personalized driving styles.

Future work will extend the proposed framework to broader traffic conditions, more complex scenarios, and richer driver preference expressions. The current framework represents each driving style using three discrete intensity levels. We can explore continuous style-intensity modeling, so that personalized driving parameters can be adjusted more smoothly and precisely according to individual user preferences.
 Moreover, driver-in-the-loop experiments will be conducted to validate whether the generated driving styles align with human subjective preferences and comfort expectations.

\appendices
% \section{Prompt-Style Description of the RAG Intent Classifier}
% \label{app:prompt_style}
% \input{Pictures/prompt}

\section{Personalized Longitudinal Planning Analysis}
\label{app:personalized_longitudinal}

Personalized longitudinal behavior during lane-change scenarios is achieved by regulating the trade-off among velocity tracking, smoothness, and responsiveness within the final speed optimization stage. 
Specifically, personalization is realized by tuning the cost-function weights of the piecewise-jerk speed optimizer, while preserving the original structure of Apollo's LaneFollow planning pipeline.

\begin{figure}[H]
\centering
\includegraphics[width=\columnwidth]{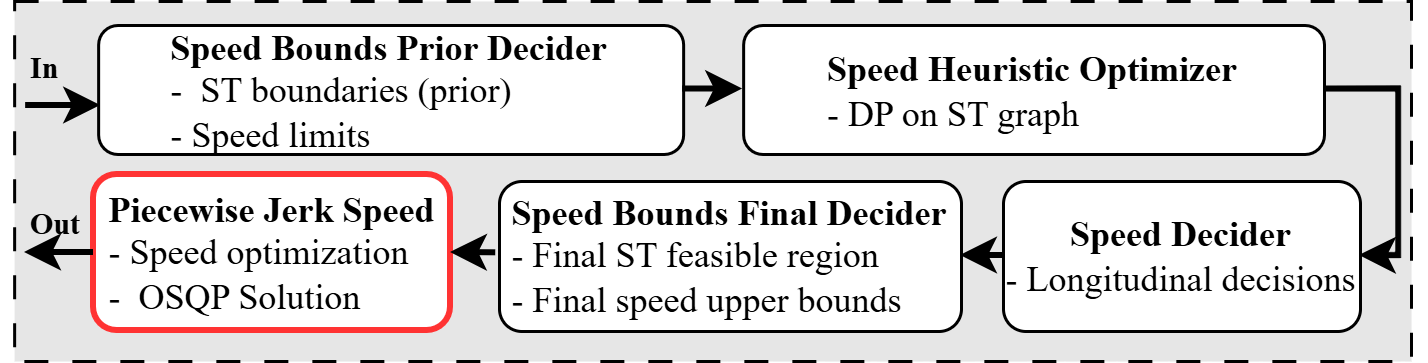}
\caption{The longitudinal speed planning process in Apollo.}
\label{fig_3}
\end{figure}

\autoref{fig_3} illustrates the longitudinal speed planning pipeline in Apollo. The pipeline first constructs spatio-temporal (ST) boundaries and obstacle-related longitudinal decisions, such as \texttt{stop}, \texttt{follow}, \texttt{yield}, and \texttt{overtake}. A coarse speed profile is generated by dynamic programming, and the final smooth speed profile is obtained by the \textit{Piecewise Jerk Speed} optimizer, which solves a constrained quadratic optimization problem using Operator Splitting Quadratic Program (OSQP) solver. In this work, the selected longitudinal parameters are mainly associated with this optimization process and are used to adjust acceleration behavior, longitudinal progress, and lane-change duration.

According to this pipeline, the piecewise-jerk speed optimizer computes the optimal trajectory within the feasible region determined by upstream modules. 
The constraints of the optimization problem can be categorized into feasibility constraints, dynamic consistency constraints, and smoothness-related constraints. The longitudinal state at time step $t_i$ is defined as $(s_i, v_i, a_i)$, where $s_i$, $v_i$, and $a_i$ denote the longitudinal position, velocity, and acceleration. 

The feasibility constraints define the admissible longitudinal motion region. 
The ST boundaries impose the positional constraint
\begin{equation}
s_{\min}(t_i) \leq s_i \leq s_{\max}(t_i),
\end{equation}
where $s_{\min}(t_i)$ and $s_{\max}(t_i)$ are determined by obstacle-induced ST boundaries after applying longitudinal decisions. 
The velocity is constrained by
\begin{equation}
0 \leq v_i \leq v_i^{\mathrm{limit}},
\end{equation} 
where $v_i^{\mathrm{limit}}$ denotes the allowable upper speed bound at time step $t_i$, which is determined by the local road speed limit, global planning speed bound, and scenario-dependent constraints. In addition, the acceleration is bounded by vehicle dynamic limits
\begin{equation}
a_{\min} \leq a_i \leq a_{\max},
\end{equation}
where $a_{\min}$ and $a_{\max}$ denote the minimum and maximum longitudinal acceleration, determined by the vehicle's physical limits. These bounds define the admissible dynamic envelope of the vehicle and indirectly influence the aggressiveness of the generated speed profile.

The dynamic consistency constraints enforce the physical relationships among these states
\begin{equation}
v_{i+1} - v_i - \frac{\Delta t}{2}(a_i + a_{i+1}) = 0,
\end{equation}
\begin{equation}
s_{i+1} - s_i - \Delta t\, v_i - \frac{\Delta t^2}{3}a_i - \frac{\Delta t^2}{6}a_{i+1} = 0,
\end{equation}
where $\Delta t = 0.1~\mathrm{s}$ is the discretization interval. 
These constraints ensure that the optimized trajectory is physically realizable.

The smoothness constraint is imposed through the jerk bound
\begin{equation}
j_{\min} \leq \frac{a_{i+1} - a_i}{\Delta t} \leq j_{\max},
\end{equation}
which limits the rate of change of acceleration and improves ride comfort and control stability.

Within this constrained feasible region, the longitudinal trajectory is obtained by solving a quadratic optimization problem. 
The objective function is formulated as
\begin{equation}
\begin{aligned}
\min_{\{s_i, v_i, a_i\}} \; J
=& \sum_{i=0}^{N-1} w_s \left(s_i - s_i^{\mathrm{ref}}\right)^2
+ \sum_{i=0}^{N-1} w_v \left(v_i - v_i^{\mathrm{ref}}\right)^2 \\
&+ \sum_{i=0}^{N-1} w_a a_i^2
+ \sum_{i=0}^{N-2} w_j \left(\frac{a_{i+1} - a_i}{\Delta t}\right)^2  .
\end{aligned}
\end{equation}

The reference position $s_i^{\mathrm{ref}}$ is obtained by temporally sampling the upstream coarse speed profile generated by the heuristic optimizer. 
The reference velocity is defined as
\begin{equation}
v_i^{\mathrm{ref}} = \min \left(v_i^{\mathrm{limit}},\, v_{\mathrm{cruise}}\right),
\end{equation}
where $v_{\mathrm{cruise}}$ is the desired cruising speed.

\begin{figure}[!b]
\centering
\includegraphics[width=\columnwidth]{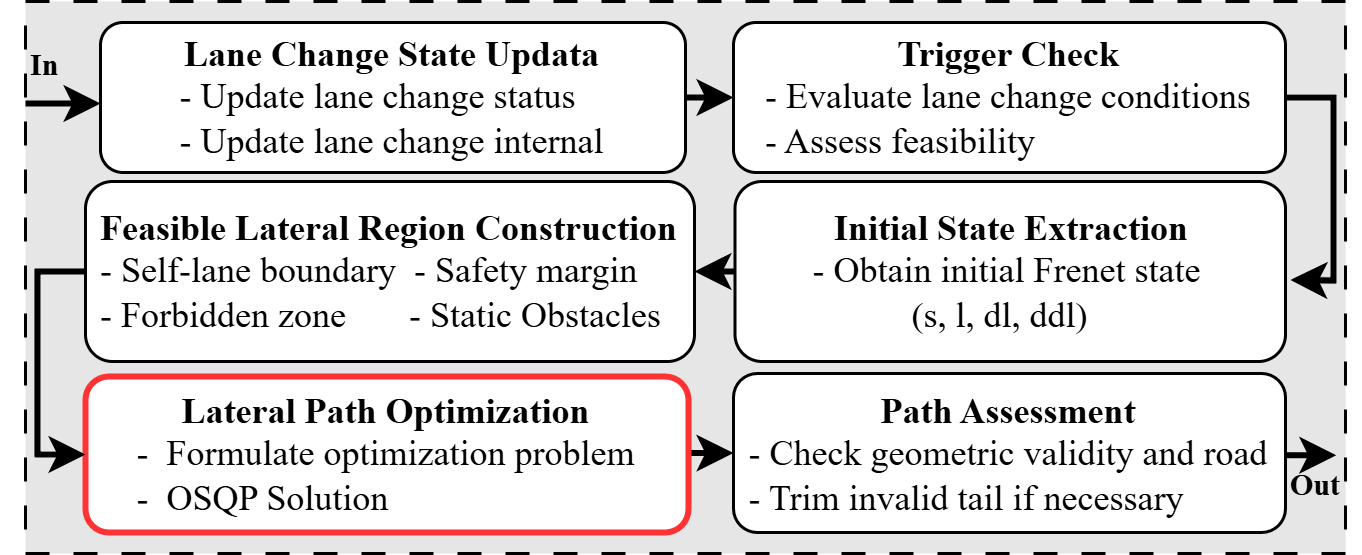}
\caption{The lane-change trajectory generation process in Apollo.}
\label{fig_4}
\end{figure}

\section{Personalized Lateral Planning Analysis}
\label{app:personalized_lateral}

To enable personalized lane-change behaviors, this work focuses on Apollo's piecewise-jerk lateral path optimizer and adjusts its geometric constraints, derivative bounds, and objective-function weights. \autoref{fig_4} illustrates the lane-change trajectory generation process in Apollo. The pipeline first updates the lane-change state and internal lane-change status through the \textit{Lane Change State Update} module. The \textit{Trigger Check} module then evaluates whether the routing-based lane-change request satisfies the required feasibility conditions. Once the request is accepted, the \textit{Initial State Extraction} module obtains the initial Frenet state, while the \textit{Feasible Lateral Region Construction} module builds the feasible lateral corridor based on self-lane boundaries, forward zones, obstacle constraints, and a vehicle-dependent safety margin. Based on these inputs, the \textit{Lateral Path Optimization} module formulates and solves a constrained quadratic optimization problem using OSQP. Finally, the generated path is checked by the \textit{Path Assessment} module to ensure geometric validity and road feasibility. In this work, the selected lateral parameters are mainly associated with the feasible lateral region, derivative bounds, and objective-function weights, and are used to adjust lane-change smoothness, lateral aggressiveness, and spatial feasibility.

In this pipeline, the personalized parameters considered are mainly associated with the feasible lateral corridor, derivative bounds, and the objective-function weights of the optimizer. The feasible lateral corridor constrains the lateral offset at each discretized station point
\begin{equation}
l_{\min}(\xi_i;\delta_{\mathrm{adc}}) \leq l_i \leq l_{\max}(\xi_i;\delta_{\mathrm{adc}}),
\end{equation}
where $\xi_i=\xi_0+i\Delta \xi$ is the discretized reference-line arc length, $l_i=l(\xi_i)$ is the lateral offset, and $\delta_{\mathrm{adc}}$ denotes the safety margin used to adjust the feasible lateral region.

In addition to the lateral corridor, the optimized path is constrained by derivative bounds
\begin{equation}
|dl_i| \leq dl_{\max},
\end{equation}
\begin{equation}
ddl_i \in [ddl_i^{\min}, ddl_i^{\max}],
\end{equation}
\begin{equation}
\frac{ddl_{i+1}-ddl_i}{\Delta \xi}
\in [dddl_i^{\min}, dddl_i^{\max}],
\end{equation}
where $dl_i=l'(\xi_i)$, $ddl_i=l''(\xi_i)$, and $(ddl_{i+1}-ddl_i)/\Delta \xi$ approximate the first-order, second-order, and third-order lateral derivatives with respect to the reference-line arc length. 

The dynamic consistency constraints enforce the geometric relationships among consecutive lateral states
\begin{equation}
dl_{i+1} - dl_i - \frac{\Delta \xi}{2}(ddl_i + ddl_{i+1}) = 0,
\end{equation}
\begin{equation}
l_{i+1} - l_i - \Delta \xi\, dl_i
- \frac{\Delta \xi^2}{3}ddl_i
- \frac{\Delta \xi^2}{6}ddl_{i+1} = 0,
\end{equation}
where $\Delta \xi$ is the discretization interval along the reference-line arc length. These constraints ensure that the optimized lateral trajectory remains geometrically consistent in the Frenet frame.

Given the above constraints, the lateral trajectory is obtained by solving a constrained quadratic optimization problem. For clarity, the main tunable terms used for personalization are written as
\begin{equation}
\begin{aligned}
\min_{\{l_i,\,dl_i,\,ddl_i\}} \; J
= &\sum_{i=0}^{N-1} w_l l_i^2
+ \sum_{i=0}^{N-1} w_{dl} dl_i^2
+ \sum_{i=0}^{N-1} w_{ddl} ddl_i^2 \\
&+ \sum_{i=0}^{N-2} w_{dddl}
\left(\frac{ddl_{i+1}-ddl_i}{\Delta \xi}\right)^2 .
\end{aligned}
\end{equation}

\bibliographystyle{IEEEtran}
\bibliography{refs}

\begin{IEEEbiography}[{\includegraphics[width=1in,height=1.25in,clip,keepaspectratio]{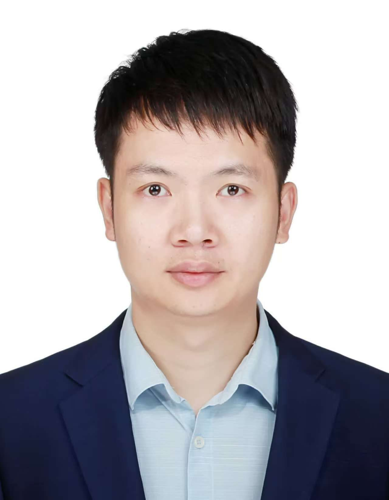}}]{Dong Bi} received the M.Sc. degree in Mechanical Engineering from Hubei University of Automotive Technology, Shiyan, China, in 2017. From 2017 to 2024, he served as a Lecturer at the same university, where he was involved in research and teaching in the areas of advanced driver assistance systems and functional safety testing for automated driving systems. Since October 2024, he has been pursuing the Ph.D. degree in the field of automated driving at Graz University of Technology, Graz, Austria. 
\end{IEEEbiography}

% \vspace{-2.0\baselineskip}

\begin{IEEEbiography}[{\includegraphics[width=1in,height=1.25in,clip,keepaspectratio]{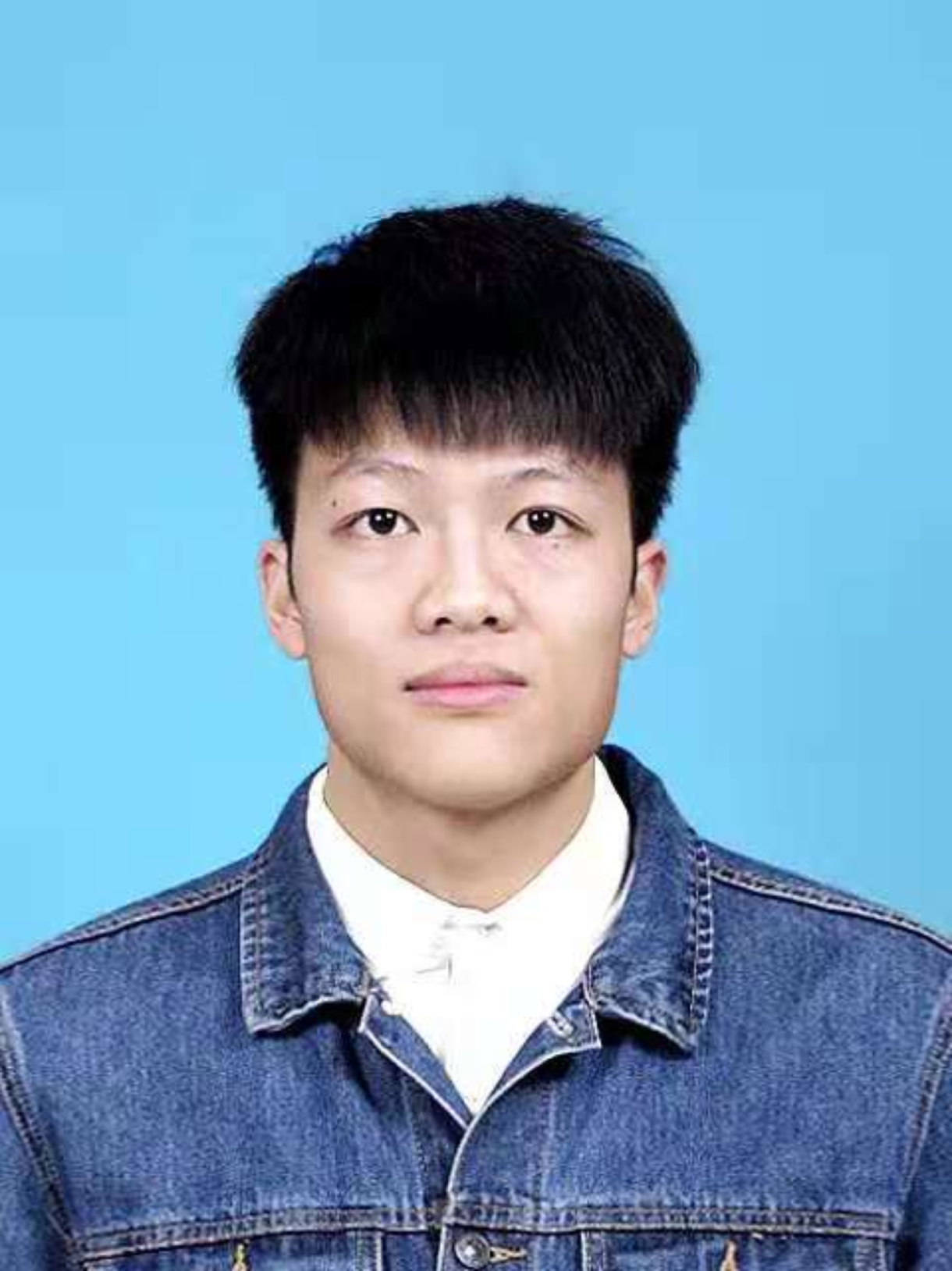}}]{Yongqi Zhao} received the bachelor’s degree from the China University of Petroleum (East China), Qingdao, China, in 2019, and the master’s degree from Technical University of Braunschweig, Braunschweig, Germany, in 2022. He is currently pursuing the Ph.D. degree with the Institute of Automotive Engineering, Graz University of Technology, Graz, Austria, with a research focus on virtual testing of automated driving systems. While pursuing his master’s degree, he gained practical experience through internships with Momenta, Stuttgart, Germany, and Volkswagen Group, Wolfsburg, Germany.
\end{IEEEbiography}
% \vspace{-2.0\baselineskip}

\begin{IEEEbiography}[{\includegraphics[width=1in,height=1.25in,clip,keepaspectratio]{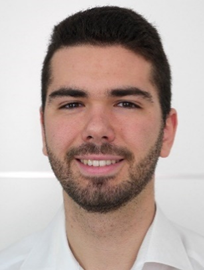}}]{Paul Kovacevic } received his bachelor's degree from Graz University of Technology, Graz, Austria, in 2023, and his master's degree from the same institution in 2025. He is currently pursuing his Ph.D. at the Institute of Automotive Engineering, where his research focuses on Cooperative Intelligent Transportation Systems (C-ITS). Alongside his academic career, he gained practical industry experience at AVL List and Magna International.
\end{IEEEbiography}
% \vspace{-2.0\baselineskip}

\begin{IEEEbiography}[{\includegraphics[width=1in,height=1.25in,clip,keepaspectratio]{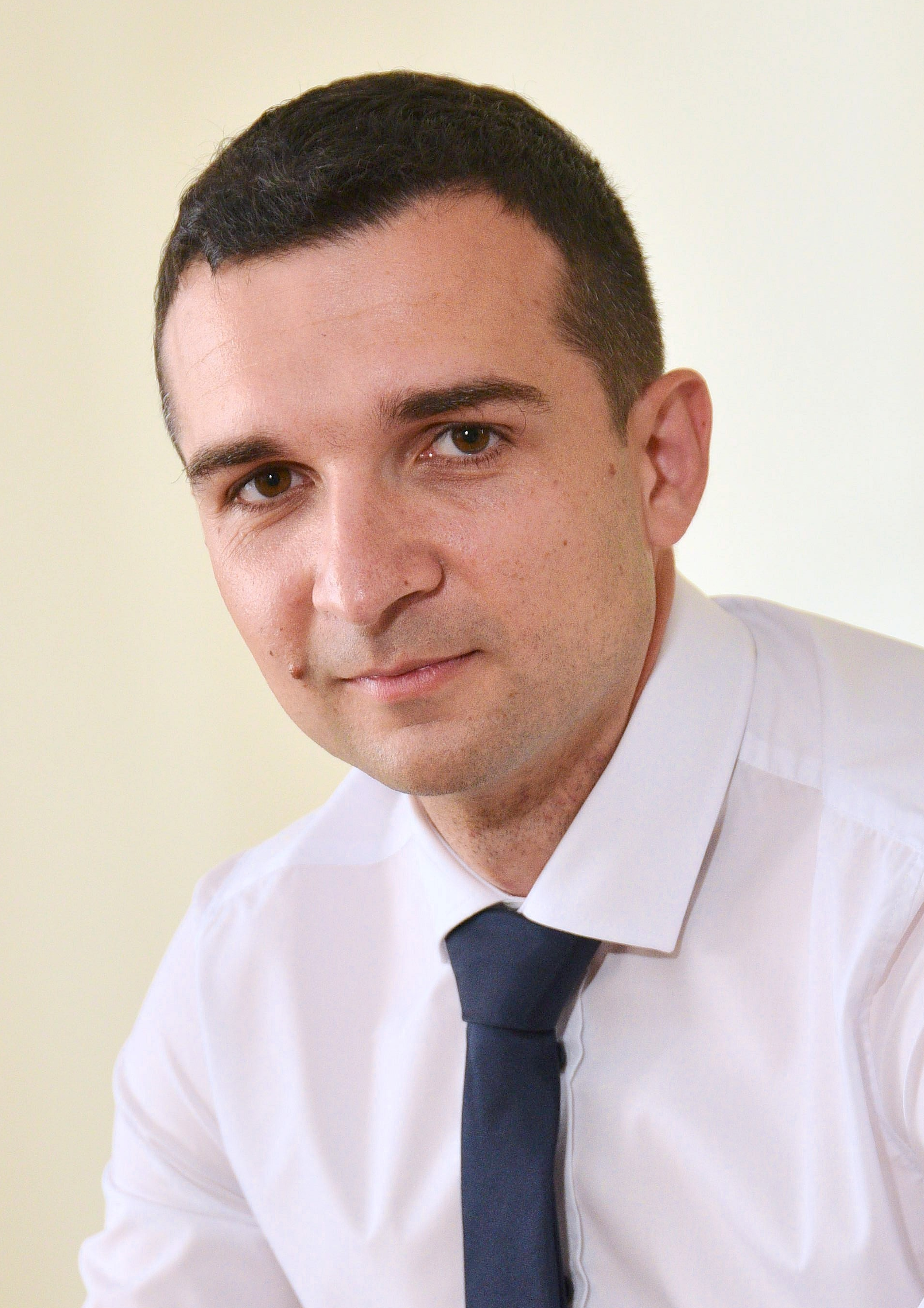}}]{Tomislav Mihalj}
received his degree in Mechanical Engineering from the University of Zagreb, Croatia, in 2014, and earned his PhD from Graz University of Technology, Austria, in 2024. From 2014 to 2019, he worked as a Research Engineer at Virtual Vehicle, Graz, Austria, where he focused on the mechanical efficiency of combustion engines, vibration analysis, and crack propagation in wheel-rail contact. Between 2019 and 2024, he served as a University Project Assistant at the Institute of Automotive Engineering, Graz University of Technology, concentrating on the virtual verification of automated driving systems. Since 2024, he has been a Postdoctoral Researcher at the same institute, where he focuses on the verification and validation of driver assistance systems and supervises related research projects.
He has authored or co-authored 12 peer-reviewed publications on rail vehicles and driver assistance systems. He has also contributed to several national and EU-funded projects related to automated driving.
\end{IEEEbiography}

% \vspace{-2.0\baselineskip}

\begin{IEEEbiography}[{\includegraphics[width=1in,height=1.25in,clip,keepaspectratio]{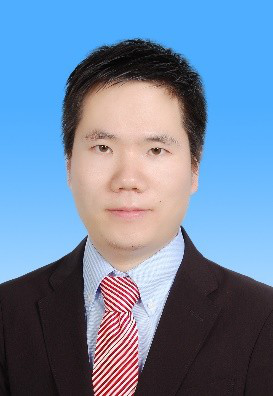}}]{Ji Zhou} 
obtained the master's degree (M.Sc) in Automotive Engineering from Jilin University, Changchun, China, in 2009. He is currently pursuing a Ph.D. degree at the Institute of Automotive Engineering, Graz University of Technology, Graz, Austria. He has worked in Ricardo, former PSA group and now in Stellantis group for more than 15 years. He has rich technical experience in powertrain control Software development, validation \& calibration, vehicle Electrical \& Electronic (EE) architecture integration validation, and EE product industrialization. His current research interests are mainly in the advanced methodology of Automated Driving Systems integration and validation. He has published 4 technical papers indexed by EI Compendex, 2 patents officially granted, and 11 other patents currently under publication.
\end{IEEEbiography}

\begin{IEEEbiography}[{\includegraphics[width=1in,height=1.25in,clip,keepaspectratio]{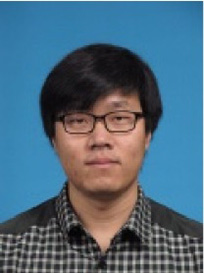}}]{Jiayuan Gong} 
received his PhD degree in Acoustics from University of Chinese Academy of Sciences, and completed postdoctoral research at Harbin Engineering University, is an associate professor of Hubei University of Automotive Technology, and the vice dean of School of Intelligent Connected Vehicle. He is a Senior Member of the China Computer Federation (CCF), serving as an Executive Member of CCF Intelligent Vehicle Committee, and a Member of the Wuhan Chapter. Additionally, he is a Committee Member of the Artificial Intelligence Division of the China Society of Automotive Engineers and a Committee Member of the Embodied Intelligence Committee of the Chinese Association for Artificial Intelligence. He leads the Electro-Electronic Information Architecture \& Vehicle-Road-Cloud Collaboration Team, and serves as Deputy Director of the Shiyan Key Laboratory of Air-Ground Swarm Collaborative Intelligence. His main research interests include CPS, HPC, Data Science, and AI.
\end{IEEEbiography}

% \vspace{-4.0\baselineskip}

\begin{IEEEbiography}[{\includegraphics[width=1in,height=1.25in,clip,keepaspectratio]{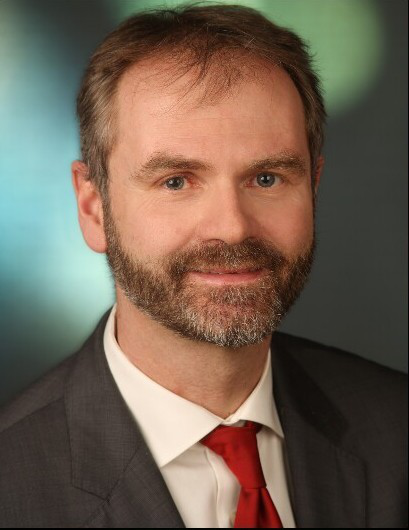}}]{Arno Eichberger} (Member, IEEE) received the degree in mechanical engineering and the Ph.D. degree (Hons.) in technical sciences from the Graz University of Technology, Graz, Austria, in 1995 and 1998, respectively. 

From 1998 to 2007, he was employed with Magna Steyr Fahrzeugtechnik AG\&Company, Graz, where he dealt with different aspects of active and passive safety. Since 2007, he has been working with the Institute of Automotive Engineering, Graz University of Technology, dealing with driver assistance systems, vehicle dynamics, and suspensions. Since 2012, he has been an Associate Professor holding a “venia docendi” of automotive engineering.
\end{IEEEbiography}

\end{document}